\documentclass[runningheads]{llncs}

\usepackage{graphicx}
\usepackage{amsmath}
\usepackage{xcolor} 
\usepackage{tabularx}
\usepackage{booktabs}
\usepackage{rotating}
\usepackage{multirow}
\usepackage{amssymb}
\usepackage{float}
\usepackage{array}
\usepackage{url}
\usepackage{scrextend}
\renewcommand{\footnotesize}{\fontsize{6pt}{9.6pt}\selectfont}

\usepackage[bookmarksnumbered,hypertexnames=false,colorlinks=true,linkcolor=blue,urlcolor=blue,citecolor=blue,anchorcolor=green,breaklinks=true]{hyperref} 
\begin{document}
\title{A Meta-Generation framework for Industrial System Generation}
%
%
\author{Fouad Oubari \inst{1,2}\and
Raphael Meunier\inst{2} \and
Rodrigue Décatoire\inst{2}\and
\\Mathilde Mougeot\inst{1}}


\institute{ Université Paris-Saclay, CNRS, ENS Paris-Saclay, Centre Borelli, Gif-sur-Yvette, France
\email{\{fouad.oubari,mathilde.mougeot\}@ens-paris-saclay.fr}\\
\and
Manufacture Française des Pneumatiques Michelin, Clermont-Ferrand, France\\
\email{\{raphael.meunier,rodrigue.decatoire\}@michelin.com}}
%
\maketitle              
\begin{abstract}

Generative design is an increasingly important tool in the industrial world. It allows the designers and engineers to easily explore vast ranges of design options, providing a cheaper and faster alternative to the trial and failure approaches. Thanks to the flexibility they offer, Deep Generative Models are gaining popularity amongst Generative Design technologies. However, developing and evaluating these models can be challenging. The field lacks accessible benchmarks, in order to evaluate and compare objectively different Deep Generative Models architectures. Moreover, vanilla Deep Generative Models appear to be unable to accurately generate multi-components industrial systems that are controlled by latent design constraints. To address these challenges, we propose an industry-inspired use case that incorporates actual industrial system characteristics. This use case can be quickly generated and used as a benchmark. We propose a Meta-VAE capable of producing multi-component industrial systems and showcase its application on the proposed use case.

\end{abstract}
\section{Introduction}
\label{introduction}

As industrial systems grow increasingly complex, the need for accurate and efficient simulation models in fields ranging from manufacturing processes \cite{mourtzis2014simulation,beniak2019research} to drug discovery \cite{liu2018molecular,durrant2011molecular} becomes ever more critical. The ability to generate new designs automatically opens up new possibilities for various domains, including topology optimization \cite{deng2022self,sosnovik2019neural}, and shape synthesis \cite{yilmaz2020conditional,brock2016generative}.
However, traditional methods for generating new prototypes suffer from inflexibility  \cite{otto1998product,pham1998parametric}, particularly when it comes to changing conception constraints. These optimization methods rely heavily on the accurate formulation of the constraints that define the object being modeled.

Deep Generative Models (DGMs) offer a flexible alternative for constrained industrial product generation. By learning the product distributions and allowing interpolation within their latent space, DGMs can generate new data with desired properties \cite{xu2019deep,oubari2021binded}. However, as observed in our experiments traditional DGMs such as Variational AutoEncoders (VAEs)\cite{kingma2013auto} and Generative Adversarial Networks (GANs) \cite{goodfellow2020generative} fall short when trying to generate more complex industrial systems, which are typically composed of multiple components connected by different latent conception constraints. Indeed, the generation process becomes challenging for mainstream DGMs when they must account for a large number of constraints and components in complex industrial systems.

In engineering, the generative design community faces a significant challenge due to the scarcity of suitable datasets and the lack of accessible benchmarks for evaluating and comparing different generative design models \cite{regenwetter2023beyond,chang2022towards}. This scarcity can be attributed to the proprietary nature of most datasets in the industrial world, leaving few open-source options available. Moreover, these limited open-source datasets may not be annotated for specific applications or suitable for generative design problems due to their complexity.

Available datasets often involve complex representations, such as voxel \cite{wu20153d} and mesh representations \cite{willis2021joinable}, which require a large amount of memory and computational resources. Additionally, industrial design frequently involves performance constraints; however, many available datasets lack performance measures or the analytic formulas necessary for computing them. This absence necessitates the approximation of the performance measure function, which can be challenging.

Other major challenges related to creating datasets for industrial design include addressing feasibility concerns \cite{greminger2020generative,cang2017microstructure} and the process-related issues. The difficulty of enforcing manufacturing constraints, leading to a discrepancy between the simulated dataset distribution and the desired one. Additionally, the inability to control training dataset properties, such as sparsity, makes it challenging to pinpoint whether a lack of performance stems from the model or the dataset. Controlling these properties allows for a better understanding of the model implementation and the dataset's effects on the model, enabling the isolation of model architecture-related and dataset-related issues. Moreover, the process of generating large and suitable datasets can be complex and time-consuming \cite{gajek2022recommendation}, often rendering traditional methods, such as computer-aided design (CAD) \cite{sarcar2008computer}, unsuitable for the task. As a result, dataset size frequently becomes a limiting factor in industrial design, highlighting the need for more efficient and effective dataset generation techniques.

The challenges related to accessing and generating suitable datasets have led to a corresponding lack of standardized benchmarks for evaluating and comparing generative design models in engineering. This hinders the development and validation of new and improved techniques in the field. As far as our knowledge extends, there is an urgent need for reliable benchmark datasets, akin to what the MNIST dataset offers for the computer vision domain as a widely-used and well-established benchmark. Such benchmarks would serve as a valuable basis for assessing and comparing generative design models in engineering. 

In order to address the limitations of current generative design models, and datasets, we present two novel contributions in this paper.{\footnotesize\footnote{The source code and dataset for this project can be found anonymously at \url{https://github.com/HiddenContributor/A-Meta-Generation-framework-for-Industrial-System-Generation}}}

\begin{itemize}
    \item Firstly, we design an industry-inspired toy dataset that is both memory-efficient and easily manipulable, thus enabling it to serve as a benchmark for evaluating the performance of various generative design models.
    \item  Secondly, we propose the Meta-VAE architecture, which is designed to jointly generate the different unitary components of a constrained industrial system. It is composed of pretrained marginal generators and a Meta-Generator that controls their generations. Notably, we show on our toy dataset that our proposed model surpasses the performance of traditional DGMs, such as the Vanilla VAE and GAN models.
\end{itemize}

\section{Achieving Balance: A Multi-Layered Industrial Use Case with Contact Constraint and Performance Measure}
\label{sec:industrial_system}
In the context of industry, the definition of a system is an industrial product that consists of one or multiple \textit{unitary components} assembled based on certain \textit{design constraints} that govern their interactions. These constraints can be implicit and apply to the unitary components as well as their interactions. A system typically comprises components of \textit{different scales} and must fulfill a specific need, which can be quantified using \textit{performance measures}. We propose hereafter a use case that captures the key attributes of an industrial system.

\begin{figure}[t]
\includegraphics[width=\textwidth]{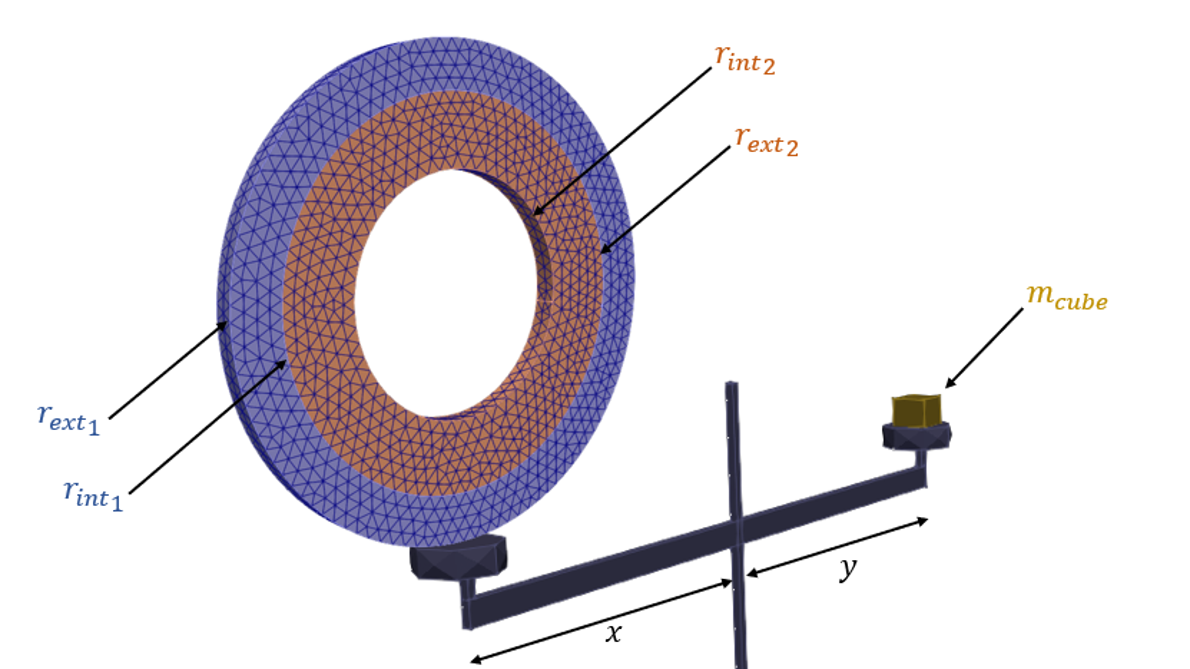}
\caption{Visual representation of the industry inspired use case.} \label{fig:use_case}
\end{figure}

We design a toy system that is an assembly of two nested cylinders, each with its own set of parameters. The goal of this assembly is to achieve balance under certain constraints, such as maintaining contact between the cylinders and verifying an equilibrium equation  as in Fig. \ref{fig:use_case}.  

We introduce the \textit{multiplicity of the unitary} components by proposing a system consisting of two hollow cylinders, $c_1$ and $c_2$, with internal radii $r_{int_1}$ and $r_{int_2}$, external radii $r_{ext_1}$ and $r_{ext_2}$.
The difference in scale is introduced by adding the densities of the cylinders' materials $d_1$ and $d_2$. 

The \textit{implicit constraint} in our use case is the following \textit{Contact Constraint}: the inner radius of the outer cylinder is equal to the external radius of the inner cylinder, i.e., $r_{{int}1} = r_{{ext}_2}$.

The use case highlights the \textit{difference in scale} between the unitary components of the system, as we consider both the shape and density of the cylinders. Finally, the use case includes a \textit{performance measure} that involves verifying the equilibrium equation between the system and a given mass $m_{cube}$, given the distances to the scale stem $x$ and $y$. The equilibrium equation can be defined as follows: 
\begin{equation}
\label{equ:equilibrium}
\begin{split}
 [\pi(r_{ext_1}^2-r_{int_1}^2)d_1+ \pi(r_{ext_2}^2-r_{int_2}^2)d_2]y = m_{cube}x.
\end{split}
\end{equation}

The constraints and the performance measure govern the interactions between the unitary components of the system and allow for the emergence of a collective behavior that is representative of an industrial system.

Despite its straightforward appearance, the use case is a simplified representation of complex industrial systems. By using a balanced, simplified assembly, we can focus on understanding the fundamental principles of such systems. This allows us to evaluate the performance of our proposed model in a controlled setting.

In this way, our use case can serve as a starting point for more complex industrial systems. It enables us to gradually build up to more realistic scenarios, which may involve a greater number of components and interactions. At the same time, we maintain an understanding of the underlying principles.

For instance, gear assemblies, robotic arms, cranes, and lifting equipment all involve multiple geometry elements, contact constraints, and the need for equilibrium verification. In gear assemblies, we must consider contact constraints and force transmission between gears. In robotic arms, components such as joints, actuators, and links interact with one another. This interaction ensures proper positioning and load-carrying capacity. Meanwhile, cranes and lifting equipment must maintain stability and balance during operation.

\section{Evaluation metrics}
\label{sec:metrics}

Our proposed use-case provides two easily computable evaluation tools: the contact constraint absolute error and the performance measure. The contact constraint absolute error is defined as:

\begin{equation}
\label{equ:contact_error}
E_c := r_{{ext}_2} - r_{{int}_1}.
\end{equation}

Here, $r_{{ext_2}}$ is the external radius of the inner cylinder and $r_{{int}_1}$ is the internal  radius of the outer cylinder.

The performance measure is defined as the distance to the system equilibrium, given by:

\begin{equation}
\label{equ:equ_error}
E_p := \pi\left[(r_{ext_1}^2-r_{int_1}^2)d_1+ (r_{ext_2}^2-r_{int_2}^2)d_2\right]y - m_{cube}x
\end{equation}

These metrics provide a way to quantify the performance of DGM in fulfilling the design constraints and performance objectives in the context of our use case.

In addition to these metrics, we compare the joint and marginal distributions of the generated and training radii and densities. To quantify the dissimilarity between the real and generated joint distributions of the parameters, we compute the Wasserstein distance \cite{villani2009wasserstein} between them.

\section{The Meta-VAE }
\label{meta_vae}
\begin{figure}[t]
\includegraphics[width=\textwidth]{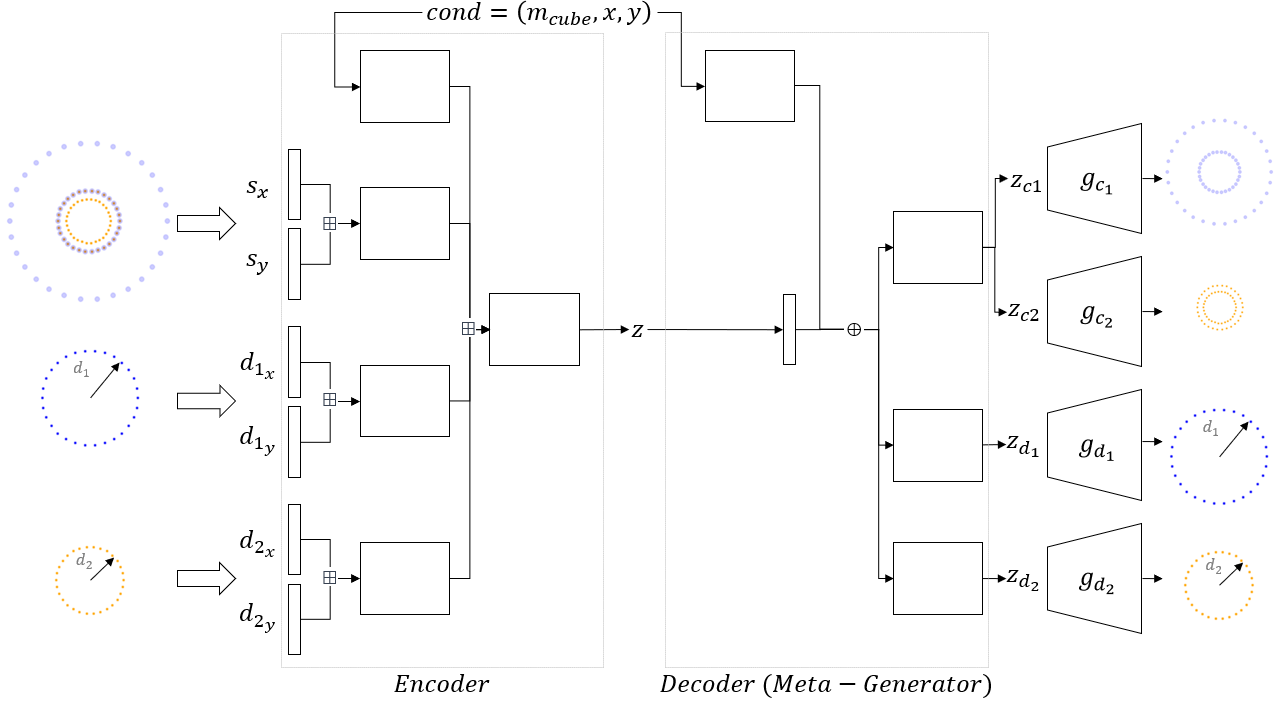}
\caption{The figure depicts the architecture the Meta-VAE applied to our use case. The marginal generators $g_i$ are pretrained. The encoder takes as input the system $s$, the densities $d_1$ and $d_2$ corresponding to each cylinder from the system, and the condition $cond = (x,y,m_{cube})$ and generates a mean $\mu$ and standard deviation $\sigma$. The decoder (Meta-Generator) takes as input a sample $z\sim \mathcal{N}(\mu,\sigma)$ and the condition $cond$ and generates latent encodings $z_i$, which are fed to the marginal generators in order to generate the unitary components $g_i(z_{c_i})$ and the densities $g_i(z_{d_i})$, resulting in a reconstruction of $s$. As detailed in the annex Sec. \ref{sec:appendix_dataset} the cylinders and densities are represented with point clouds. The system $s$ is the concatenation of the two cylinders and is expressed as the set of the coordinates $(s_x,s_y)$. The densities are expressed as sets of coordinates $(d_{i_x},d_{i_y})_{i\in\{ 1,2 \}}$. The blocks in this figure consist of fully connected and ReLU layers. The symbols $\boxplus$ and $\oplus$ represent the concatenation and summation operations, respectively.}
\label{fig:meta_vae}
\end{figure}

In this section, we introduce the Meta-VAE, a novel model designed to generate accurate industrial systems. The Meta-VAE is composed of a Meta-Generator and marginal generators, which work together to produce a system that meets the required conditions (see Fig. \ref{fig:meta_vae}). The Meta-Generator generates latent encodings for each marginal generator, corresponding to the correct part of a system that satisfies all the given constraints and performance measures for a given condition $cond=(m_{cube},x,y)$ (refer to Sec. \ref{sec:industrial_system}).

Separating the blocks inside the encoder and decoder enables us to balance the importance given to each unitary component by controlling the depth of each block and thus provides better results than a single encoding and decoding fully connected block. 

In a general setting, the Meta-VAE is trained using a Vanilla conditional VAE approach. In the encoding phase, the encoder takes as input the system $s=(x_1,\cdots,x_n)$ a condition $cond$ where $x_i$ are the unitary components. The condition $cond$ could correspond in an industrial setting to a conception desired property. The encoder then outputs the mean $\mu$ and standard deviation $\sigma$ of the encoded distribution. In the decoding phase, using the encoded mean and standard deviation, a posterior $z_{meta} = \mu + \sigma\odot \epsilon$ , $\epsilon\sim \mathcal{N}(0,I)$ is sampled and fed to the decoder alongside the condition $cond$. The Meta-Decoder outputs the latent vectors $z_i$ that are fed to each marginal generator $g_i$. This creates a correlation between the marginal generations $g_i(z_i)$ in order to obtain unitary components that can be assembled into a correct system, optimized with respect to the condition $cond$ alongside other conditions . 

The training loss for the Meta-VAE is a modified version of the VAE loss:

\begin{equation}
\label{equ:meta_vae_loss}
    \begin{split}
        \mathcal{L}_{\text{Meta-VAE}} &= \sum_{i=1}^n \text{MSE}(x_i,g_i(z_i)) + \text{KL}(\mathcal{N}(\mu,\sigma),\mathcal{N}(0,I)),
    \end{split}
\end{equation}
where \text{KL} is the Kullback-Leibler divergence, and \text{MSE} the mean squared error.

In our setting, the components $\{x_i\}$ correspond to the outer and inner cylinders $c_1$ and $c_2$, as well as their respective densities $d_1$ and $d_2$ (as described in Sec. \ref{fig:use_case}).

\section{Related work}
\label{sec:related_work}

While some datasets have been developed to train Deep Generative Models for design tasks \cite{sosnovik2019neural,willis2021fusion}, few of them include multi-component products, which are common in industrial design. For example, McComb et al. \cite{mccomb2018data} introduced a tabular truss design dataset containing $1379$ truss structure designs. This dataset encompasses sequential design operations such as joint and structural member placement, as well as performance metrics like safety factors and weights. Another example is the work of  Regenwetter et al. \cite{regenwetter2022biked}, they propose a dataset of $4500$ bike designs generated by multiple designers using BikeCAD software, which includes different assembly and component images, numerical design parameters, and class labels.

Several studies investigate the application of sequential approaches in generative design. Raina et al. \cite{raina2019learning} developed a framework based on Reinforcement Learning \cite{kaelbling1996reinforcement} (RL) and utilizing deep learning constructs, imitation learning, and image processing to extract human design strategies from the truss dataset \cite{regenwetter2022biked}. The approach imitates human strategies by observing design state sequences, enabling the \text{RL} agent to generate competitive truss designs without explicit guidance. Wu et al. \cite{wu2021deepcad} propose a transformer-based model to generate sequential operations that can be viewed as the drawing process of CAD models shape creation. CAD representations can also provide hierarchical relationships that can represented as graphs, which opens some possibilities for deep generative design. For their experiments, Gajek et al. \cite{gajek2022recommendation} also employed CAD catalogues. In their paper, the authors propose an approach that recommends the next component for an assembly by treating mechanical assemblies as undirected graphs and training graph neural networks GNN \cite{wu2020comprehensive}. This allows them to predict the next component given a partial graph. Some papers explore different multi-components systems representations for their deep generative design  methods.

Other works have investigate alternative multi-component system representations for deep generative design methods. Stump et al \cite{stump2019spatial} use a grammar-based representation to defined systems and propose a method to optimize both the system's form and behaviour. They use  RL  in a physics-based simulated environment to incorporate their generated system's behaviour to train an RNN to generate the semantics of high performing designs.

More straightforward generative approaches remain hard to develop, mainly due to the unavailability of benchmarks in this domain. This challenge makes it difficult to compare the performance of different generative models on a benchmark dataset. Our proposed dataset can serve as benchmark for evaluating the performance of different Deep Generative Models for industrial system generation.

In this work, we also propose an architecture that builds upon state-of-the-art Deep Generative Models to generate complex industrial systems with multiple components related by different physical constraints. Our proposed architecture overcomes the limitations of existing Deep Generative Models for joint distribution generation and provides improved results as shown is Sec. \ref{sec:experiments}. By proposing an architecture that leverages state-of-the-art models as marginal generators, we contribute to the advancement of deep generative design and provide a solid foundation for further research in this domain.

\section{Experiments}
\label{sec:experiments}
We conduct our experiment using custom datasets and compare our framework against vanilla DGM presented in the following section.

\subsection{Dataset Representation}
\label{sec:dataset_generation}
To represent the cylinders in our system, we use point clouds to represent their internal and external radii. The densities are represented by circular point clouds, where the density value corresponds to the circle radius. To train the models, we use a dataset of size $20000$. For more details, refer to the appendix Sec.\ref{sec:appendix_dataset}

\subsection{Baselines}	
\label{sec:baseline}

\subsubsection{Marginals Generators}
\label{sec:marginal_def}

To generate the marginal distributions, we train Variational AutoEncoders on cylinders and density distributions. These VAEs are trained to generate point cloud representations of the densities and the cylinders. Once trained, their decoders are used as marginal generators in the Meta-VAE.

\subsubsection{Meta Generators}
\label{sec: meta_gen_def}

In this study, we evaluate several generative models, including a vanilla conditional VAE, a vanilla conditional GAN, and the Meta-VAE. In addition, we introduce a simplified version of the Meta-VAE, which we refer to as the Simplified Meta-VAE (SMVAE) . The SMVAE does not include the marginal generators present in the original Meta-VAE, instead it directly generates the unitary components. Both the Meta-VAE and SMVAE are composed of parallel encoder and decoder blocks. The vanilla models are trained on whole systems, which are the concatenations of the nested cylinders and their corresponding densities. All models comprise fully connected layers and ReLU activations.
To ensure the robustness of our results, we train each model five times and use the generated outputs from all five models for our analysis and figures presented in Sec. \ref{sec:results}.  For simplicity, we will use the terms `Vanilla VAE’ and `Vanilla GAN’ to refer to the vanilla conditional VAE and GAN models, respectively, in the subsequent sections.

\subsection{Results and discussion}
\label{sec:results}
\subsubsection{Generations Visualization}

\begin{figure}[t!]
\includegraphics[width=\textwidth]{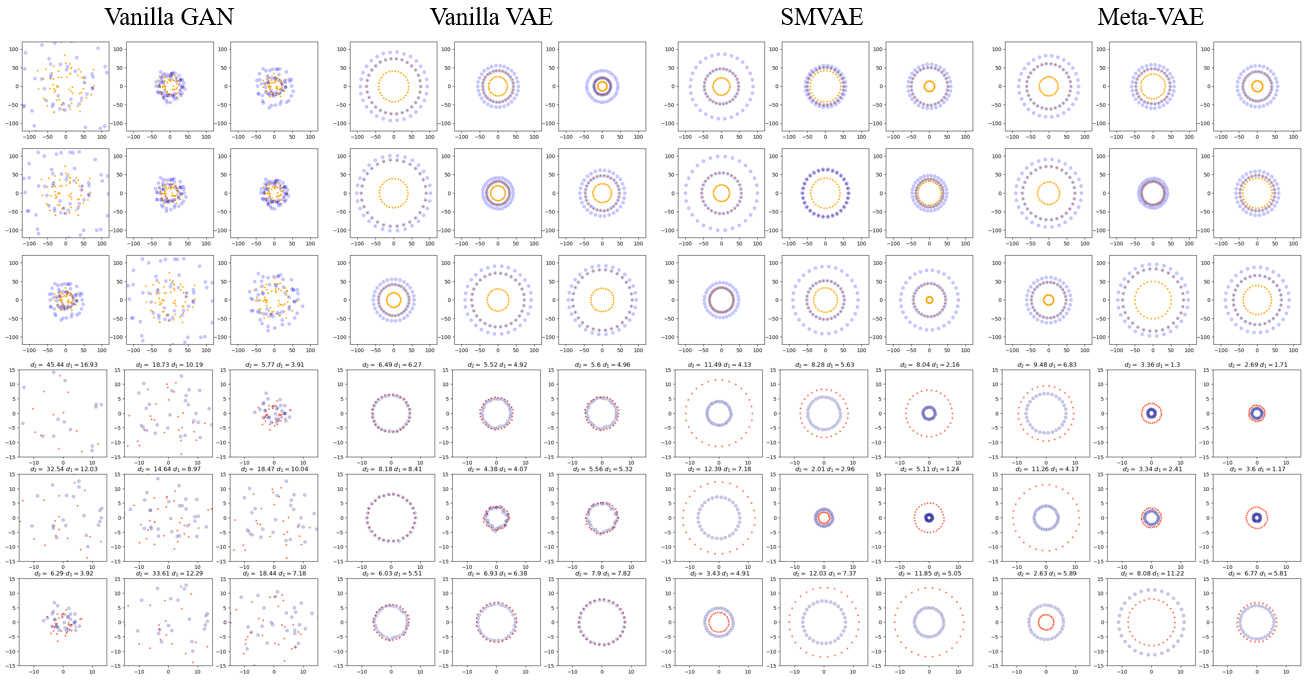}
\caption{Scatter plots depicting the generated cylinders and densities for the four generative models: Vanilla GAN, Vanilla VAE, Simplified Meta-VAE (SMVAE), and Meta-VAE. The outer cylinder point clouds are shown in blue, and the inner cylinder point clouds are shown in orange. The outer cylinder's density $d_1$ is shown in red, and the inner cylinder's density $d_2$ in blue.}
\label{fig:generations}
\end{figure}

Visual inspection of the generated systems produced by the four generative models reveals that the Simplified Meta-VAE (SMVAE) and the Meta-VAE outperform the Vanilla GAN and Vanilla VAE models (Fig. \ref{fig:generations}). Notably, the systems generated by the SMVAE and Meta-VAE appear to respect the contact constraint, as further explored in Sec. \ref{sec:constraints_results}. Conversely, the Vanilla GAN models produce unsatisfactory results in terms of both the variability and quality of the generated systems.

While the systems generated by the Vanilla VAE model appear to be correct, the density generations are not accurate and suffer from a lack of variability. This lack of variability suggests that the equilibrium constraint might not be fully met. Overall,  the observations imply that the SMVAE and Meta-VAE models may exhibit better performance than the vanilla models in generating realistic and diverse systems.

\subsubsection{Distributions comparison} \textcolor{white}{.}
\label{sec:distrib_comparison}

\begin{figure}[t!]
\includegraphics[width=\textwidth]{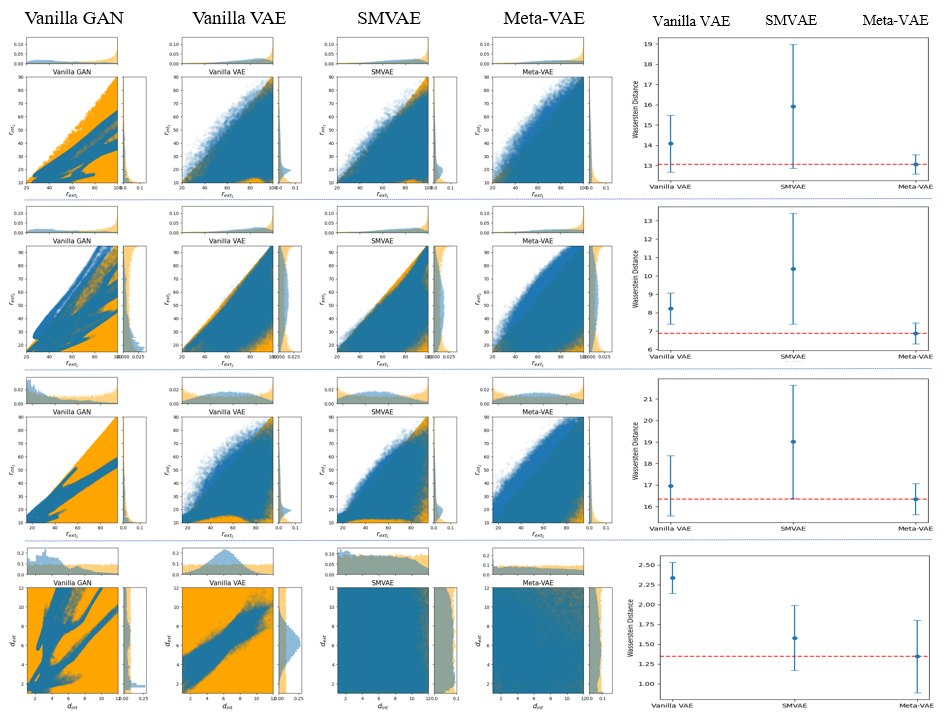}
\caption{Joint and marginal distributions of generated systems' radii and densities and their corresponding Wasserstein distance error plots. Generated distributions are depicted in blue, while real distributions are shown in orange, both corresponding to the same set of conditions $cond=(x,y,m_{cube})$. The first three rows represent specific pairs of radii joint distributions: $(r_{ext_1}, r_{int_2})$, $(r_{ext_1}, r_{ext_2})$, and $(r_{ext_2}, r_{int_2})$. The last row represents the densities $d_1$ and $d_2$ joint distributions. Columns, from left to right, represent the Vanilla GAN, Vanilla VAE, SMVAE, and Meta-VAE models. Wasserstein error plots display the mean and variance of the Wasserstein distances between the real and generated joint distributions for each model, based on five training runs. Due to the significantly larger means and standard deviations of the Vanilla GAN's Wasserstein distances, its error plots have been omitted to enable clearer comparison among the other models.}
\label{fig:joint_distr}
\end{figure}

\textbf{- Variability:} In order to check the generation variability, we plot the joint distributions of the generated systems' radii and densities (Fig.\ref{fig:joint_distr}).

We conduct further analysis on the generated systems by comparing the joint distributions of the internal  and external radii of both the inner and outer cylinders of the generated systems with those of the real systems. Our results indicate that the Meta-VAE exhibits the best variability amongst all models, covering the entire support of the joint distribution. The SMVAE and Vanilla VAE seem perform well, covering similar proportions of the real distribution support.  However, upon examining the marginal distributions, we observe a lack of variability in the generated internal  radii of the inner cylinder $r_{i_{2}}$. The Vanilla GAN shows the poorest performance in terms of system variability.

With regards to the joint distributions of the densities, the Vanilla VAE exhibits some variability, but displays a tendency towards generating only a few values of $r_{{int}_{i_{i\in \{ 1,2 \}}}}$ for each value of $r_{{ext}_{i_{i\in\{ 1,2 \}}}}$, and vice versa. Although this tendency is not precisely linear, it is sufficiently pronounced to suggest that the models generate relatively few distinct pairs of internal  and external radii values. On the other hand, the Vanilla GAN performs the worst out of the three models, failing to generate distributions that are close to the real distribution. The Meta-VAE and SMVAE produce the most variability by covering the entire support of the real joint distribution.
\newpage

\begin{figure}[t!]
\includegraphics[width=\textwidth]{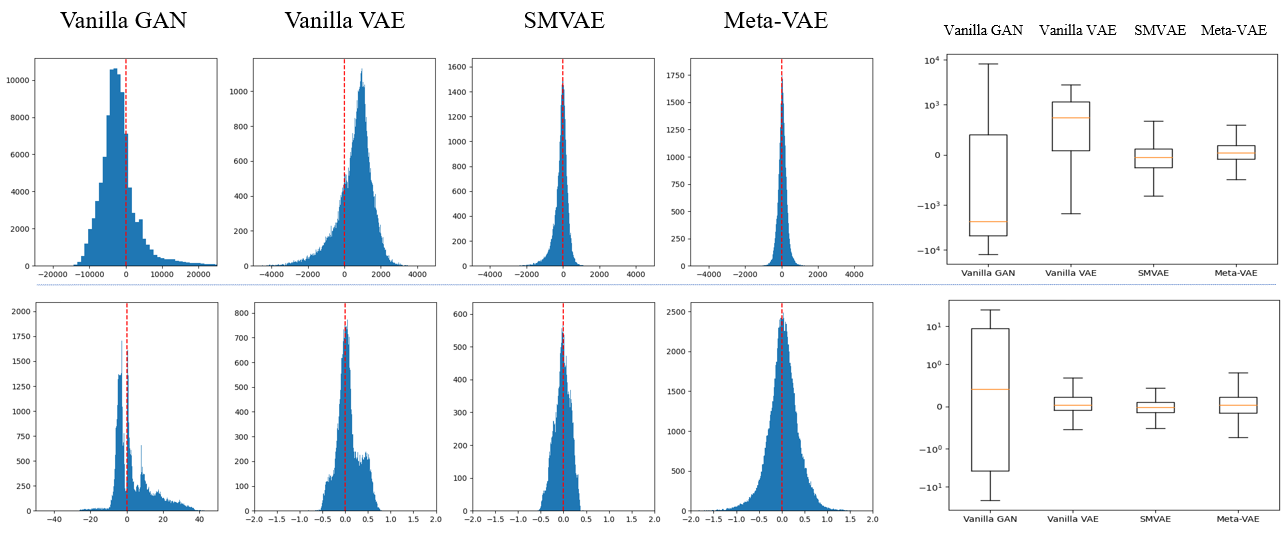}
\caption{Histograms and box plots of the \textit{equilibrium constraint} absolute error $E_p$ (top) and \textit{contact condition} absolute error $E_c$ (bottom) for the four Deep Generative Models used in our experiments. Sample size: $5\times 10^4$.} \label{fig:abs_errors}
\end{figure}

\textbf{ - Distance to Real Distribution:} To assess the capacity of each model to accurately reproduce the real joint distribution, we compute the Wasserstein distance between the real and generated joint distributions of radii and densities is calculated. The Wasserstein density values (Fig.\ref{fig:abs_errors}) collectively indicate that the Meta-VAE surpasses the performance of other models. It is noteworthy that the radii distribution generated by the Vanilla VAE more closely approximates the real distribution compared to that of the SMVAE, while the opposite holds true for the densities distributions.

\textbf{- Synthesis:} These observations  suggest that the Meta-VAE's architecture effectively manages the unitary components' multiplicity by adeptly coordinating the marginal generators, a feature less pronounced in the other models, particularly the Vanilla VAE. Consequently, these alternative models encounter greater challenges in striking a balance between the unitary components, resulting in diminished performance in comparison

\subsubsection{Contact constraint and equilibrium verification}\textcolor{white}{.}
\label{sec:constraints_results}

\begin{figure}[t!]
\includegraphics[width=\textwidth]{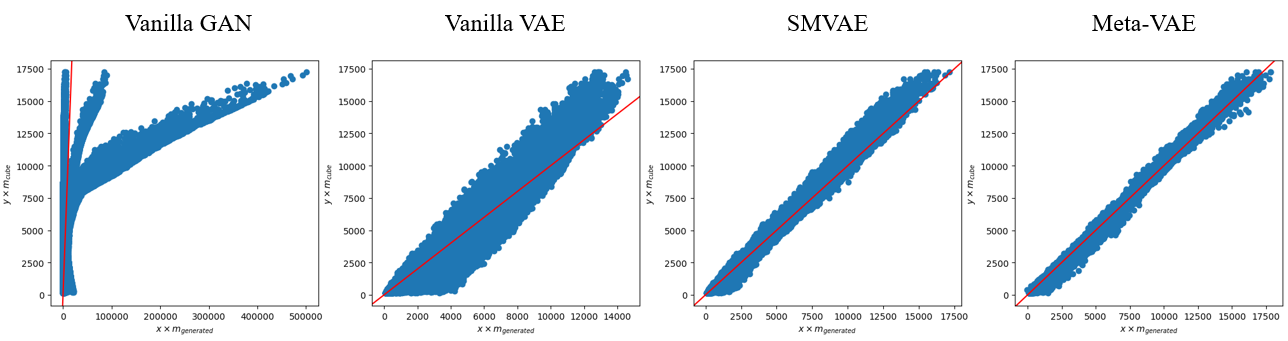}
\caption{Residual plots showing the agreement between the two terms of the performance measure $m_{generated}\times y$ and $m{cube}\times x$ (Eq.\ref{equ:equilibrium}) for the Vanilla GAN  (left), the Vanilla VAE (middle), and the Meta-VAE (right). A perfect model would produce a line with a slope of 1 and an intercept of 0.  Sample size: $5\times10^4$.} \label{fig:equilibrium_pairwize}
\end{figure}

We evaluate the ability of the generated systems and densities to verify the equilibrium equation, as outlined in Sec. \ref{sec:industrial_system}, by computing the performance measure absolute error $E_p$ (Eq. \ref{equ:equ_error}). The histograms and corresponding box plots of the performance absolute error $E_p$ (Fig. \ref{fig:abs_errors}) provide insight into the model's capacity to satisfy the equilibrium constraint. The Meta-VAE demonstrates the strongest performance in effectively meeting this constraint, with the SMVAE following closely behind. While the Vanilla VAE exhibits some limitations in accurately satisfying the equilibrium constraint, as indicated by a larger mean and standard deviation, it still performs better than the Vanilla GAN, which displays the weakest performance in this regard, characterized by the largest mean error and standard deviation. The residual plots (Fig. \ref{fig:equilibrium_pairwize}) confirm this observation, as the Vanilla VAE errors are not as closely aligned with the bisector compared to those of the Meta-VAE and the SMVAE, while the Vanilla GAN errors show a substantial misalignment, indicating a much poorer performance in adhering to the equilibrium constraint.

Similarly, to evaluate the verification of the contact constraint, we compute the contact constraint absolute error (Eq. \ref{equ:contact_error}). Fig. \ref{fig:abs_errors} reveals that the contact constraint is satisfactorily enforced by the Vanilla VAE, SMVAE, and Meta-VAE. In contrast, the Vanilla GAN displays a substantially larger mean error and standard deviation, indicating its considerably poorer performance in enforcing the contact constraint compared to the other models.

Overall, the results suggest that the Meta-VAE outperforms the other models in terms of simultaneously and accurately enforcing both the equilibrium and the contact constraints, an aspect where the vanilla models struggle. This superior performance could be attributed to the fact that the marginal generators in the Meta-VAE handle the distribution learning of the unitary components, enabling the Meta-Generator to focus solely on learning the correct joint distribution and capturing the implicit constraint. Although the SMVAE, Vanilla VAE, and Vanilla GAN do not have marginal generators and are trained to generate the entire system of unitary components as a whole, the SMVAE exhibits performance comparable to that of the Meta-VAE, which can be attributed to their similar architectures. Thus, the proposed architecture appears to effectively reduce the complexity of the learning task for the Meta-Generator, allowing it to focus on learning the correct joint distribution while the marginal generators handle the distribution learning of the unitary components.

\subsubsection{Overall comparison}\textcolor{white}{.}
\label{sec:overall}
The Meta-VAE outperforms all other models in terms of both quality and variability of the generated designs. Specifically, the Meta-VAE generates designs that better verify the equilibrium equation than the other models, and its generations also perform comparatively well in terms of verifying the contact constraint. In contrast, while the generations of the other models seem to verify the contact constraint better, they exhibit a lack of variability, particularly in the marginal distributions of the internal  radius of the inner cylinder.

The multiplicity of unitary components and the latent constraint and performance measures make the generation process challenging for standard Deep Generative Models. For instance, the Vanilla VAE can generate seemingly correct cylinders but cannot generate correct densities and fails to correctly enforce the verification of the equilibrium on its generations. The Vanilla GAN performs worse than the other models, introducing an additional challenge of achieving model convergence. The lack of convergence in the GAN model results in a significant degradation in the quality of the generated designs, making the generative design task even more difficult. 

We infer that the key to the Meta-VAE's outperformance is the division of the generation task into a marginal and a joint generation. By training marginal generators to generate the system's unitary components, the Meta-VAE can focus solely on the joint distribution, allowing it to enforce the latent contact constraint and perform better with respect to the performance measure. This separation enables the Meta-VAE to accurately catch the system constraints and focus on the ``assembling" task of a system design.

Furthermore, we believe that the Meta-VAE architecture can scale well with respect to the difficulty and complexity of the industrial systems to be generated. As discussed, the Meta-VAE is better at capturing the correlations between the marginal distributions, making it suitable for more complex generation tasks. In such scenarios, instead of training a single large model, the Meta-Generator offers the flexibility of using pre-trained models for marginal generation. This can potentially save computation resources and time that would otherwise be required for training a large model on complex systems. Thus, the Meta-VAE can be considered a contribution to the generative design field, as it enables building upon existing state-of-the-art models, which can be used as marginal generators.

\section{Conclusion}
\label{sec:conclusion}
The application of Deep Generative Models in generative design for industrial systems is still in its early stages, with traditional DGM facing difficulties in generating complex systems. In this work, we introduced the Meta-VAE, a novel architecture that demonstrates superior performance in terms of quality and variability of generated designs, as showcased by our balanced simplified nested cylinder assembly. Building upon these promising results, we believe that the Meta-VAE architecture has a potential to scale effectively with the complexity of industrial systems to be generated. The model's ability to focus on joint distribution and enforce latent constraints, as well as its flexibility in utilizing pre-trained models for marginal generation, offers significant advantages in terms of computational efficiency and adaptability. Indeed, our findings suggest that utilizing pretrained models as marginal generators does not result in any degradation of performance.
Our proposed use case encapsulates the essential characteristics of real-world industrial systems while being easy to generate and manipulate, making it a valuable benchmark for the generative design community.
 By showcasing the practical relevance of our use case, we aim to inspire further research and development. We encourage others to adopt and build upon our work, ultimately fostering the creation of additional benchmarks in a domain where they are currently scarce. In this dynamic landscape, as the application of Deep Generative Models in generative design for industrial systems continues to evolve, we believe that our model can play a valuable role in advancing this field.

\section{Ethical Statement}
Despite our primary focus on addressing engineering challenges, we acknowledge the ethical implications of our research and the possibility of unintended consequences or misuse in certain applications :
\begin{itemize}
    \item Responsible Use of Generated Systems:
    We encourage the responsible use of our model and the generated systems in alignment with ethical guidelines, industry best practices, and regulations.
    \item Dual-use Concerns:
    Our research is intended for peaceful and non-harmful applications. We urge users of our model and generated systems to consider the ethical implications of their use in different contexts.
    \item Transparency and Openness:
    We promote transparency and openness by making our work, including datasets and models, available to the research community to encourage collaboration, replication, and further development.
\end{itemize}
\bibliographystyle{unsrt}
\typeout{}
\bibliography{bibliography}

\begin{thebibliography}{10}

\bibitem{mourtzis2014simulation}
Dimitris Mourtzis, Michael Doukas, and Dimitra Bernidaki.
\newblock Simulation in manufacturing: Review and challenges.
\newblock {\em Procedia Cirp}, 25:213--229, 2014.

\bibitem{beniak2019research}
Juraj Beniak, Michal Holdy, Peter Kri{\v{z}}an, and Milo{\v{s}}
  Mat{\'u}{\v{s}}.
\newblock Research on parameters optimization for the additive manufacturing
  process.
\newblock {\em Transportation Research Procedia}, 40:144--149, 2019.

\bibitem{liu2018molecular}
Xuewei Liu, Danfeng Shi, Shuangyan Zhou, Hongli Liu, Huanxiang Liu, and Xiaojun
  Yao.
\newblock Molecular dynamics simulations and novel drug discovery.
\newblock {\em Expert opinion on drug discovery}, 13(1):23--37, 2018.

\bibitem{durrant2011molecular}
Jacob~D Durrant and J~Andrew McCammon.
\newblock Molecular dynamics simulations and drug discovery.
\newblock {\em BMC biology}, 9(1):1--9, 2011.

\bibitem{deng2022self}
Changyu Deng, Yizhou Wang, Can Qin, Yun Fu, and Wei Lu.
\newblock Self-directed online machine learning for topology optimization.
\newblock {\em Nature communications}, 13(1):388, 2022.

\bibitem{sosnovik2019neural}
Ivan Sosnovik and Ivan Oseledets.
\newblock Neural networks for topology optimization.
\newblock {\em Russian Journal of Numerical Analysis and Mathematical
  Modelling}, 34(4):215--223, 2019.

\bibitem{yilmaz2020conditional}
Emre Yilmaz and Brian German.
\newblock Conditional generative adversarial network framework for airfoil
  inverse design.
\newblock In {\em AIAA aviation 2020 forum}, page 3185, 2020.

\bibitem{brock2016generative}
Andrew Brock, Theodore Lim, James~M Ritchie, and Nick Weston.
\newblock Generative and discriminative voxel modeling with convolutional
  neural networks.
\newblock {\em arXiv preprint arXiv:1608.04236}, 2016.

\bibitem{otto1998product}
Kevin~N Otto and Kristin~L Wood.
\newblock Product evolution: a reverse engineering and redesign methodology.
\newblock {\em Research in engineering design}, 10:226--243, 1998.

\bibitem{pham1998parametric}
DT~Pham.
\newblock Parametric and feature-based cad/cam concepts, techniques,
  applications by jj shah and m. m{\"a}ntyl{\"a}, wiley, chichester, 1995, 619
  pp., isbn 0--471--00214--3 ({\pounds} 55; hbk).
\newblock {\em Robotica}, 16(6):701--702, 1998.

\bibitem{xu2019deep}
Youjun Xu, Kangjie Lin, Shiwei Wang, Lei Wang, Chenjing Cai, Chen Song, Luhua
  Lai, and Jianfeng Pei.
\newblock Deep learning for molecular generation.
\newblock {\em Future medicinal chemistry}, 11(6):567--597, 2019.

\bibitem{oubari2021binded}
Fouad Oubari, Antoine De~Mathelin, Rodrigue D{\'e}catoire, and Mathilde
  Mougeot.
\newblock A binded vae for inorganic material generation.
\newblock {\em NeurIPS 2021 Workshop on Deep Generative Models and Downstream
  Applications}, 2021.

\bibitem{kingma2013auto}
Diederik~P Kingma and Max Welling.
\newblock Auto-encoding variational bayes.
\newblock {\em arXiv preprint arXiv:1312.6114}, 2013.

\bibitem{goodfellow2020generative}
Ian Goodfellow, Jean Pouget-Abadie, Mehdi Mirza, Bing Xu, David Warde-Farley,
  Sherjil Ozair, Aaron Courville, and Yoshua Bengio.
\newblock Generative adversarial networks.
\newblock {\em Communications of the ACM}, 63(11):139--144, 2020.

\bibitem{regenwetter2023beyond}
Lyle Regenwetter, Akash Srivastava, Dan Gutfreund, and Faez Ahmed.
\newblock Beyond statistical similarity: Rethinking metrics for deep generative
  models in engineering design.
\newblock {\em arXiv preprint arXiv:2302.02913}, 2023.

\bibitem{chang2022towards}
Rees Chang, Yu-Xiong Wang, and Elif Ertekin.
\newblock Towards overcoming data scarcity in materials science: unifying
  models and datasets with a mixture of experts framework.
\newblock {\em npj Computational Materials}, 8(1):242, 2022.

\bibitem{wu20153d}
Zhirong Wu, Shuran Song, Aditya Khosla, Fisher Yu, Linguang Zhang, Xiaoou Tang,
  and Jianxiong Xiao.
\newblock 3d shapenets: A deep representation for volumetric shapes.
\newblock In {\em Proceedings of the IEEE conference on computer vision and
  pattern recognition}, pages 1912--1920, 2015.

\bibitem{willis2021joinable}
Karl~DD Willis, Pradeep~Kumar Jayaraman, Hang Chu, Yunsheng Tian, Yifei Li,
  Daniele Grandi, Aditya Sanghi, Linh Tran, Joseph~G Lambourne, Armando
  Solar-Lezama, and Wojciech Matusik.
\newblock Joinable: Learning bottom-up assembly of parametric cad joints.
\newblock {\em arXiv preprint arXiv:2111.12772}, 2021.

\bibitem{greminger2020generative}
Michael Greminger.
\newblock Generative adversarial networks with synthetic training data for
  enforcing manufacturing constraints on topology optimization.
\newblock In {\em International Design Engineering Technical Conferences and
  Computers and Information in Engineering Conference}, volume 84003, page
  V11AT11A005. American Society of Mechanical Engineers, 2020.

\bibitem{cang2017microstructure}
Ruijin Cang, Yaopengxiao Xu, Shaohua Chen, Yongming Liu, Yang Jiao, and Max
  Yi~Ren.
\newblock Microstructure representation and reconstruction of heterogeneous
  materials via deep belief network for computational material design.
\newblock {\em Journal of Mechanical Design}, 139(7):071404, 2017.

\bibitem{gajek2022recommendation}
Carola Gajek, Alexander Schiendorfer, and Wolfgang Reif.
\newblock A recommendation system for cad assembly modeling based on graph
  neural networks.
\newblock In {\em European Conference on Machine Learning and Principles and
  Practice of Knowledge Discovery in Databases (ECMLPKDD 2022)}, 2022.

\bibitem{sarcar2008computer}
MMM Sarcar, K~Mallikarjuna Rao, and K~Lalit Narayan.
\newblock {\em Computer aided design and manufacturing}.
\newblock PHI Learning Pvt. Ltd., 2008.

\bibitem{villani2009wasserstein}
C{\'e}dric Villani and C{\'e}dric Villani.
\newblock The wasserstein distances.
\newblock {\em Optimal Transport: Old and New}, pages 93--111, 2009.

\bibitem{willis2021fusion}
Karl~DD Willis, Yewen Pu, Jieliang Luo, Hang Chu, Tao Du, Joseph~G Lambourne,
  Armando Solar-Lezama, and Wojciech Matusik.
\newblock Fusion 360 gallery: A dataset and environment for programmatic cad
  construction from human design sequences.
\newblock {\em ACM Transactions on Graphics (TOG)}, 40(4):1--24, 2021.

\bibitem{mccomb2018data}
Christopher McComb, Jonathan Cagan, and Kenneth Kotovsky.
\newblock Data on the design of truss structures by teams of engineering
  students.
\newblock {\em Data in brief}, 18:160--163, 2018.

\bibitem{regenwetter2022biked}
Lyle Regenwetter, Brent Curry, and Faez Ahmed.
\newblock Biked: A dataset for computational bicycle design with machine
  learning benchmarks.
\newblock {\em Journal of Mechanical Design}, 144(3), 2022.

\bibitem{raina2019learning}
Ayush Raina, Christopher McComb, and Jonathan Cagan.
\newblock Learning to design from humans: Imitating human designers through
  deep learning.
\newblock {\em Journal of Mechanical Design}, 141(11), 2019.

\bibitem{kaelbling1996reinforcement}
Leslie~Pack Kaelbling, Michael~L Littman, and Andrew~W Moore.
\newblock Reinforcement learning: A survey.
\newblock {\em Journal of artificial intelligence research}, 4:237--285, 1996.

\bibitem{wu2021deepcad}
Rundi Wu, Chang Xiao, and Changxi Zheng.
\newblock Deepcad: A deep generative network for computer-aided design models.
\newblock In {\em Proceedings of the IEEE/CVF International Conference on
  Computer Vision}, pages 6772--6782, 2021.

\bibitem{wu2020comprehensive}
Zonghan Wu, Shirui Pan, Fengwen Chen, Guodong Long, Chengqi Zhang, and S~Yu
  Philip.
\newblock A comprehensive survey on graph neural networks.
\newblock {\em IEEE transactions on neural networks and learning systems},
  32(1):4--24, 2020.

\bibitem{stump2019spatial}
Gary~M Stump, Simon~W Miller, Michael~A Yukish, Timothy~W Simpson, and Conrad
  Tucker.
\newblock Spatial grammar-based recurrent neural network for design form and
  behavior optimization.
\newblock {\em Journal of Mechanical Design}, 141(12), 2019.

\end{thebibliography}






\newpage

\section{Appendix}
\label{sec:appendix}

\subsection{Wasserstein distance}
\begin{definition}[Wasserstein distance]
\label{def:wasserstein}
 Let $(M, d)$ be a metric space, and let $p \in [1, \infty)$. Given two probability measures $P$ and $Q$ defined on $M$, the $p$-Wasserstein distance between $P$ and $Q$ is defined as:

\begin{equation}
\label{equ:wasserstein_distance_general}
W_p(P,Q) = \left(\inf_{\gamma \in \Gamma(P,Q)} \int_{M \times M} d(x,y)^p \mathrm{d}\gamma(x,y)\right)^{\frac{1}{p}},
\end{equation}

where $\Gamma(P,Q)$ denotes the set of all joint probability measures on $M \times M$ with marginals $P$ and $Q$. 
\end{definition}
The Wasserstein distance can be seen as the optimal cost of transporting mass from one probability measure to another, with the cost measured by the product of the distance $d(x,y)$ and the amount of mass transported between $x$ and $y$.

In our study, we use on the 1-Wasserstein distance, also known as the Earth Mover's Distance, which corresponds to $p=1$. We approximate the 1-Wasserstein distance by first computing histograms for the real and generated distributions of the radii and densities, and then calculating the distance between the histograms  $H_1$ and $H_2$ as follows:

\begin{equation}
\label{equ:wasserstein_distance_histogram}
W_p(H_1,H_2) = \sum_{i=1}^M \sum_{j=1}^N  |H_1(i,j) - H_2(i,j)|
\end{equation}

where $H_1(i,j)$ and $H_2(i,j)$ are the probabilities associated with the bins $(i, j)$ of the histograms $H_1$ and $H_2$, and $M$ and $N$ are the numbers of bins along each axis of the histograms.

The Wasserstein distance provides a natural way to compare the joint distributions of the generated and real parameters in our problem. By computing the Wasserstein distance between the joint distributions of the real and generated parameters, we can quantify the dissimilarity between the distributions and evaluate the performance of the different types of models.

\subsection{Dataset generation}
\label{sec:appendix_dataset}

This section describes the procedure used to generate a dataset for the use case considered. Firstly, the use case parameters are generated following a specific procedure detailed in Sec. \ref{sec:use_case_params}. Next, we generate the representations of the cylinders using point clouds, with the densities represented using circles as explained in Sec. \ref{sec:use_case_representation}.

\subsubsection{Use case parameters}\textcolor{white}{.}
\label{sec:use_case_params}

\begin{figure}[!ht]
\label{fig:dataset_sample}
\includegraphics[width=\textwidth]{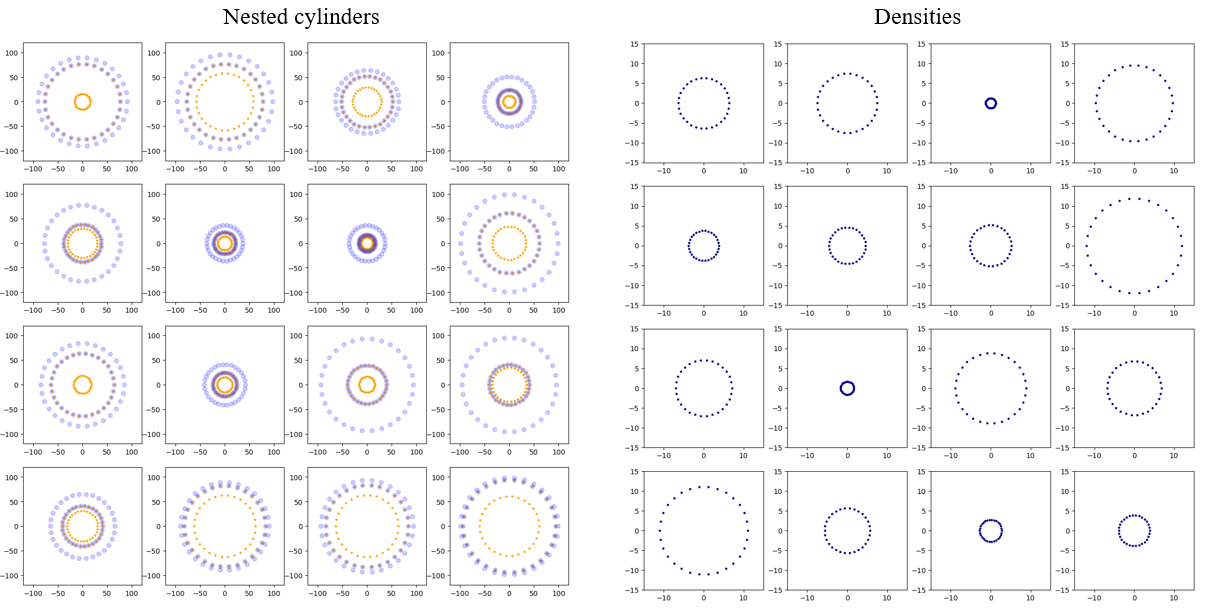}
\caption{Samples from the toy dataset.} 
\end{figure}

We generate the parameters for our use case as follows:
\begin{itemize}
\item \textbf{Cylinders generation}. We generate the radii for our cylinders by splitting the total number of radii into three equal parts. For each part, we start by generating one of the three radii ($r_{{ext_1}}$, $r_{{int_1}}$, or $r_{{int_2}}$) uniformly at random from a range of values (as described below). We then generate the other two radii conditionally, based on the first radii. Specifically:
\begin{itemize}
\item When we generate $r_{{ext_1}}$, we then generate $r_{{int_1}} = r_{{ext_2}}$ such that $r_{{ext_2}}$ is uniformly random from the range $[15,r_{{ext_1}}-t]$, and then $r_{{int_2}}$ is uniformly random from the range $[10,r_{{ext_2}}-t]$.
\item When we generate $r_{{int_1}} = r_{{ext_2}}$, we generate $r_{{ext_1}}$ uniformly random from the range $[r_{{int_1}}+t,r_{{ext_2}}-t]$ and $r_{{int_2}}$ uniformly random from the range $[10,r_{{ext_2}}-t]$.
\item When we generate $r_{{int_2}}$, we then generate $r_{{int_1}} = r_{{ext_2}}$ such that $r_{{ext_2}}$ is uniformly random from the range $[r_{{int_2}}+t,95]$, and then $r_{{ext_1}}$ is uniformly random from the range $[r_{{int_1}}+t,100]$.
\end{itemize}
\item \textbf{Density generation}. We generate the densities $d_1$ and $d_2$ uniformly at random from the range $[1,12]$.
\item \textbf{Distance generation}. We generate the distances $x$ and $y$ by generating $x$ uniformly at random from the range $[1,99]$ and setting $y=100-x$.
\item \textbf{Mass calculation}. We calculate $m_{cube}$ using the equilibrium equation (Eq. \ref{equ:equilibrium}).
\end{itemize}
Where $t$, the cylinders' thickness is set to 5.

\subsubsection{Use case representation}\textcolor{white}{.}
\label{sec:use_case_representation}

We represent the internal  and external borders of the cylinders using point clouds for simplicity and low memory usage. To generate the circular borders, we choose an angular partition $(\theta_i)_{i\in[0,29]}$ of $\left[0, +2\pi\right[$, which gives us 30 points corresponding to $\rho_i\theta_i$, where $\rho_i$ is one of the four use case radii.

We experimented with different methods to represent the densities (e.g., one hot encodings), but given the satisfactory results obtained when generating the cylinder borders using Variational AutoEncoders, we decided to adopt a similar representation. We represent the densities using circles represented as point clouds, with the corresponding density as the circle radius. Fig. \ref{fig:dataset_sample} presents samples from the dataset.

Regarding the use case hyperparameters $(x,y,m_{cube})$, we simply represent them as scalar values.

\subsection{Models architectures}
\label{sec:appendix_architectures}

\begin{figure}[!ht]
\includegraphics[width=\textwidth]{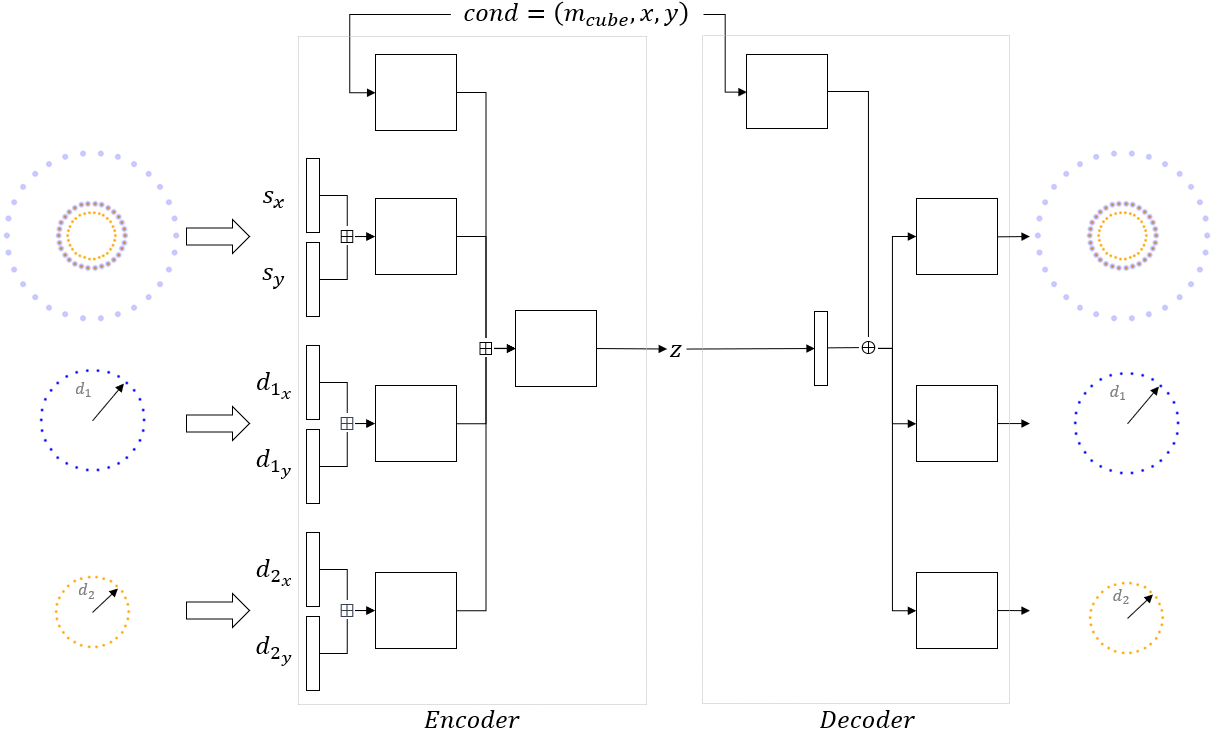}
\caption{Representation of Simplified Meta-VAE (SMVAE) and the Meta-VAE. The SMVAE architecture, the decoder features parallel blocks, each directly generating the unitary components. The model is trained using the same loss function as the Meta-VAE. The smaller boxes represent single fully connected layers with ReLU activation. The symbols $\boxplus$ and $\oplus$ indicate 
 concatenation and summation operations, respectively.} \label{fig:SMVAE}
\end{figure}

The architecture of the SMVAE is represented in figure Fig.\ref{fig:SMVAE}.

\end{document}